\documentclass[conference]{IEEEtran}

\IEEEoverridecommandlockouts
\usepackage{cite}
\usepackage{amsmath,amssymb,amsfonts}
\usepackage{algorithmic}
\usepackage{graphicx}
\usepackage{textcomp}
\usepackage{xcolor}
\usepackage{booktabs}
\usepackage{todonotes}
\usepackage[english]{babel}
\usepackage{lipsum}
\usepackage{float}
\usepackage{multirow}
\usepackage{graphicx}
\usepackage{hyperref}

\def\BibTeX{{\rm B\kern-.05em{\sc i\kern-.025em b}\kern-.08em
    T\kern-.1667em\lower.7ex\hbox{E}\kern-.125emX}}

\graphicspath{{./images/}}

\usepackage{blindtext}
\usepackage{eso-pic}

\begin{document}
\title{\vspace*{1cm}Improving Image Tracing with Convolutional Autoencoders by High-Pass Filter Preprocessing}

\author{\IEEEauthorblockN{Zineddine Bettouche and Andreas Fischer}
\IEEEauthorblockA{Deggendorf Institute of Technology \\
Dieter-Görlitz-Platz 1\\
94469 Deggendorf \\
E-Mail: \href{mailto:zineddine.bettouche@th-deg.de}{zineddine.bettouche@th-deg.de}, \href{mailto:andreas.fischer@th-deg.de}{andreas.fischer@th-deg.de}}

}

\maketitle

\begin{abstract}
The process of transforming a raster image into a vector representation is known as image tracing. This study looks into several processing methods that include high-pass filtering, autoencoding, and vectorization to extract an abstract representation of an image. According to the findings, rebuilding an image with autoencoders, high-pass filtering it, and then vectorizing it can represent the image more abstractly while increasing the effectiveness of the vectorization process. 

\end{abstract}

\begin{IEEEkeywords}
   \textit{image quality; vector graphics;  principal component analysis; neural networks; autoencoders; high-pass filters; vectorization; complexity theory; and information technology.}
\end{IEEEkeywords}

\section{Introduction}
Object recognition is considered a complex task in the processing field. Its complexity far exceeds simple arithmetic operations. With the massive amount of data generated each year, manual calculations done by hand are completely ignored. Therefore, data processing and evaluation are automated for all operations.

In recent years, many studies have emerged to contribute to the advancement of knowledge in the field of object recognition. Two of the pillars of this field are image processing and artificial intelligence (AI). AI is a fascinating subject that has attracted a lot of attention in the last decade, especially with its use in computer vision. Now not only filter-based models, e.g., Haar Cascade, can be trained to classify images, but also neural networks can be wired to learn how to detect various shapes and objects. The models generally learn from the pixel values and model their structures in mathematical equations, which begs the question of whether it would be more efficient for the models to learn from vector images as they are closer to the nature of the trained models than spatial data in the form of pixel arrays. Thus, this article is an attempt to improve the tracing of images by using autoencoders and high-pass filters to obtain an abstract representation of images in vector form. The highpass filters are chosen since they emphasize the important features of an image. This work is considered a step forward to achieving a better training rate in object recognition with ANN.

This paper is an extended version of the previously published paper~\cite{bettouche}that discussed the summarized content of the findings that this work produced. A more in-depth discussion about the techniques used in the work, such as Image Tracing, Potrace, and autoencoders, has been added in the background section. The previous papers that touched on the topic of image tracing have been further discussed in detail to illustrate the place our work takes in relation to what has already been accomplished in the field. Concerning the methodology followed, a detailed description of the autoencoding network built is provided, and the choice of the layers is justified. When it comes to the experimentation part, other experiments are added, such as the attempt at reducing the noise without blurring the images. The experiments already introduced in the previous paper are extended to further discuss their findings, and detailed images that visualize those findings are added.

In other words, concerning the added value of this paper over its previous conference version, it can be stated that every section has gone through many further details, to present a richer methodology section (as for the ease of future building over our findings), to underline the networks built, and technologies used (such as the trained autoencoders that were described layer by layer and Potrace as a vectorization tool), and to provide an extended experimentation section, as the experiments’ discussions are lengthened, detailed more with visualization of their results, and assisted with other experiments (such as a blur-free noise reduction attempt).

At first, there was the question: if the autoencoding of an image can improve its vectorized format by reconstructing its important features, how can high-pass filters come into play in the process? In other words, "Can a high-pass filter be used in combination with an autoencoding model to achieve an abstract representation of the image through the process of vectorization?" Thus, various ideas branched from this node, leading to the different pipelines that can be built to experiment with high-pass filter integration. For instance, the filters can be put before the autoencoding stage of a model that is already trained with filtered images to better reconstruct the significant data, leading to better vectorization. More systematically, the autoencoding stage can act as a smoothing process, removing the noise from the images while reducing their complexity, while the filters come afterward to further enhance the quality of the important features, leading to a more abstract representation.

The remainder of this paper is structured as follows: In Section II, an introduction is given to image tracing, autoencoders, and high-pass filters. Section III discusses related work. Section IV introduces the methodology of this paper, including the evaluation methods used and the reasons why they were chosen. Section V presents the experiments and their results. This is the part that attempts to eliminate inefficient processing algorithms so that only a few pipelines that score closely are put forward for further evaluation. Section VI includes the evaluation of the different processing pipelines built and closes with a summarizing interpretation. Finally, Section VII concludes the paper and discusses future work.

\section{Background}
\label{background}

\subsection{Image Tracing}
Image tracing is the process of converting a bitmap into a vector graphic. As Selinger writes in his tracing algorithm~\cite{selinger}, vector graphics are described as algebraic formulas of the contours, typically in the form of Bezier curves. The advantage of displaying an image as a vector outline is that it can be scaled to any size without loss of quality. They are independent of the resolution and are used, for example, for fonts, since these must be available in many different sizes. However, most input and output devices, such as scanners, displays, and printers, generate bitmaps or raster data. For this reason, a conversion between the two formats is necessary. Converting a vector graphic into a bitmap is called "rendering" or "rasterizing." Tracing algorithms are inherently imperfect because many possible vector outlines can represent the same bitmap. Of the many possible vector representations that can result in a particular bitmap, some are more plausible or aesthetically pleasing than others. For example, to render bitmaps with a high resolution, each black pixel is represented as a precise square that creates staircase patterns. However, spikes are neither pleasant to look at nor are they particularly plausible interpretations of the original image. Bezier curves are used to represent the outlines.
\begin{figure}
    \centering
    \includegraphics[width=0.2\textwidth]{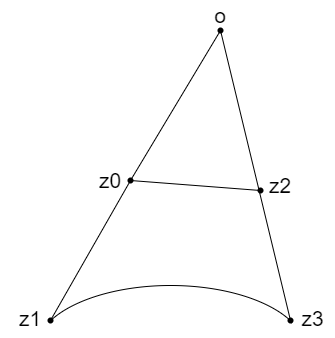}
	\caption{Bezier curve}    
	\label{fig:bezier}
\end{figure}

\begin{figure}
    \centering
    \includegraphics[width=0.49\textwidth]{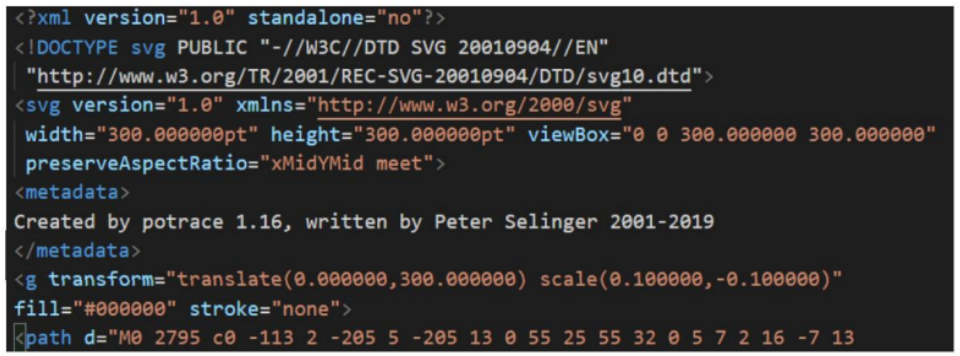}
	\caption{Header of an SVG file by potrace}    
	\label{fig:svg-file}
\end{figure}

As seen in Figure~\ref{fig:bezier}, a cubic Bezier curve consists of four control points, which determine the curvature of the curve. As a rule, the vector graphics are saved as SVG files (Scalable Vector Graphics). This file format is a special form of an XML file. XML stands for Extensible Markup Language. It is used to present hierarchically structured data in a human-readable format. As can be seen from Figure~\ref{fig:svg-file}, the structure of this file is based on the Extensible Markup Language scheme. The file header defines which versions of XML and SVG are used. The height and width of the graphic in points are also specified. In this case, the g element represents the drawing area on which to draw. The elements to be drawn consist of tags stored as XML elements. They are particularly important in connection with the path elements. Quadratic and cubic Bezier curves, as well as elliptical arcs and lines, can be put together as best fits. The entries here determine which form the path takes.

\subsection{Potrace as a Vectorization Tool}
\begin{figure}
    \centering
    \includegraphics[width=0.3\textwidth]{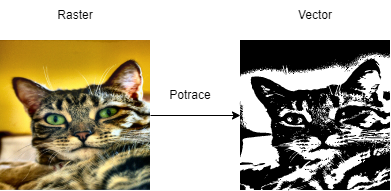}
	\caption{Potrace vectorization}    
	\label{fig:potracevectorization}
\end{figure}

Potrace is a tracing algorithm that was developed by Peter Selinger~\cite{selinger}. It is considered simple and efficient as it produces excellent results. Potrace stands for "polygon tracer," where the output of the algorithm is not a polygon but a contour made of Bezier curves. This algorithm works particularly well for high-resolution images. Potrace generates grayscale images as a threshold vector rather than as the output. The conversion from a bitmap to a vector graphic is done in several steps. First, the bitmap is broken down into several paths that form the boundaries between black and white areas. The points adjoining four pixels are given integer coordinates. These points are saved as vertices when the four adjacent pixels are not the same color. The connection between two vertices is called the edge. A path is thus a sequence of vertices, whereby the edges must all be different. The path composition in Potrace works by moving along the edges between the pixels. Every time a corner is found, a decision is made as to which direction the path will continue based on the colors of the surrounding pixels. If a closed path is defined, it is removed from the image by inverting all pixel colors inside the path. This will define a new bitmap on which the algorithm will be applied recursively until there are no more black pixels. Then its optimal polygon is approximately determined for each path. The criterion for optimality with Potrace is the number of segments. A polygon with a few segments is therefore more optimal than one with several segments. In the last phase, the polygons obtained are converted into a smooth vector outline. Here, the vertices are first corrected so that they correspond as closely as possible to the original bitmap. Furthermore, in the main step, the corners and curves are calculated based on the length of the adjacent segments and the angles between them. Optionally, the curves can be optimized after this process so that they match the original bitmap as closely as possible. Then, in the main step, the corners and curves are calculated based on the length of the adjacent segments and the angles between them. Optionally, the curves can be optimized after this process so that they match the original bitmap even more closely. Then, in the main step, the corners and curves are calculated based on the length of the adjacent segments and the angles between them. Finally, the curves can be optimized after this process. Figure~\ref{fig:potracevectorization} shows the output vector image when applying Potrace to an input raster image.

\subsection{Autoencoder}
\begin{figure}
    \centering
    \includegraphics[width=0.8\linewidth]{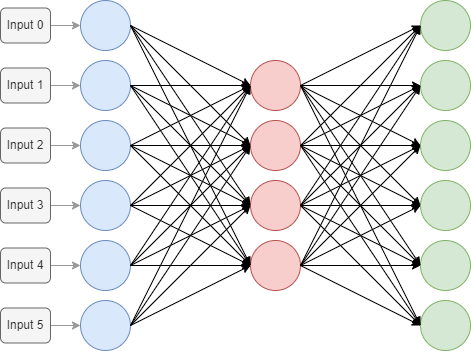}
	\caption{Example structure of an autoencoding network}    
	\label{fig:autoencoder}
\end{figure}

A typical use of a neural network is for supervised learning. It involves training data, which contains an output label. The neural network tries to learn the mapping from the given input to the given output label. Nevertheless, if the input vector itself replaces the output label, then the network will try to find the mapping from the input to itself. This would be the identity function, which is a trivial mapping. However, if the network is not allowed to simply copy the input, then it will be forced to capture only the salient features. This constraint opens up a different field of applications for neural networks, which was unknown. The primary applications are dimensionality reduction and specific data compression. The network is first trained on the given input. The network attempts to reconstruct the given input from the features it has picked up and outputs an approximation of the input. The training step involves the computation of the error and backpropagating the error. The typical architecture of an autoencoder resembles a bottleneck. Figure~\ref{fig:autoencoder} depicts the schematic structure of an autoencoder.

The encoder part of the network is used for encoding and sometimes even for data compression purposes, although it is not very effective as compared to other general compression techniques like JPEG. Encoding is achieved by the encoder part of the network, which has a decreasing number of hidden units in each layer. Thus, this part is forced to pick up only the most significant and representative features of the data. The second half of the network performs the decoding function. This part has an increasing number of hidden units in each layer and thus tries to reconstruct the original input from the encoded data. Therefore, autoencoders are an unsupervised learning technique. Training an autoencoder for data compression: For a data compression procedure, the most important aspect of compression is the reliability of the reconstruction of the compressed data. This requirement dictates the structure of the autoencoder as a bottleneck.
\begin{enumerate}
\item \textbf{Encoding the input data:} The autoencoder first tries to encode the data using the initialized weights and biases.
\item \textbf{Decoding the input data:} The autoencoder tries to reconstruct the original input from the encoded data to test the reliability of the encoding.
\item \textbf{Backpropagating the error:} After the reconstruction, the loss function is computed to determine the reliability of the encoding. The error generated is backpropagated. The above-described training process is reiterated several times until an acceptable level of reconstruction is reached.
\end{enumerate}

After the training process, only the encoder part of the autoencoder is retained to encode a similar type of data used in the training process. The different ways to constrain the network are:
\begin{itemize}
\item \textbf{Keep small Hidden Layers:} If the size of each hidden layer is kept as small as possible, then the network will be forced to pick up only the representative features of the data thus encoding the data.
\item \textbf{Regularization:} In this method, a loss term is added to the cost function which encourages the network to train in ways other than copying the input.
\item \textbf{Denoising:} Another way of constraining the network is to add noise to the input and teach the network how to remove the noise from the data.
\item \textbf{Tuning the Activation Functions:} This method involves changing the activation functions of various nodes so that a majority of the nodes are dormant thus effectively reducing the size of the hidden layers.
\end{itemize}

\subsection{High-pass Filters}
\begin{figure}
    \centering
    \includegraphics[width=0.8\linewidth]{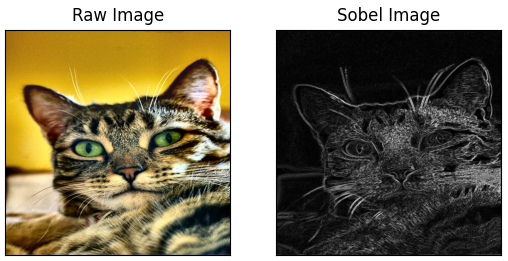}
	\caption{Applying Sobel derivatives on a random image}    
	\label{fig:sobel}
\end{figure}

A high-pass filter can be used to make an image appear sharper. These filters (e.g., Sobel~\cite{sobel} and Canny~\cite{canny}) emphasize fine details in the image. The change in intensity is used by high-pass filtering. If one pixel is brighter than its immediate neighbors, it gets boosted. Figure~\ref{fig:sobel} shows the result of applying a high-pass filter (Sobel) on a random image.

\section{Related Work}
\label{related-work}
Image segmentation can be considered an extension of image classification where localization succeeds the classification process. It is a superset of image classification with the model pinpointing where a corresponding object is present by outlining the object's boundary.  Image segmentation techniques can be divided into two classes: 
\begin{itemize}
	\item Classical computer vision approaches: such as thresholding, edge, region- or cluster-based segmentation techniques.
	\item AI-based approaches using mainly autoencoders. For instance, DeepLab made use of convolutions to replace simple pooling operations and prevent significant information loss while downsampling. 
\end{itemize}
In our paper, we focus on the use of high-pass filters with autoencoders, which succeeded with a vectorization process. Hence, the relevant work on these topics is introduced in this section.

To create better vectorize vectors, Lu et al.~\cite{lu} leverage additional depth information stored in RGB-D images. Although they anticipate consumer gear will soon be able to produce photos with depth information, this still has to happen. The method described here, however, operates with standard RGB photos without the need for additional gear.

Bera~\cite{bera} offers a different method for image vectorization. It emphasizes the advancement made possible by edge detection techniques. This study, in contrast, looks into the benefits of dimensionality reduction.

A method for vector pictures based on splines rather than Bézier curves is presented by Chen et al.~\cite{chen} To create a combination of raster and vector graphics, they concentrate on data structures that facilitate real-time editing.

Solomon and Bessmeltsev~\cite{solomon} investigated the usage of frame fields in an MIT study. Finding a smooth frame field on the image plane with at least one direction aligned with neighboring drawing outlines is the basic goal of their method. The two directions of the field will line up with the two intersecting contours at X- or T-shaped junctions. The frame field is then traced, and the traced curves are then grouped into strokes to extract the drawing's topology. Finally, they produced a vectorization that was in line with the frame field using the extracted topology.

Lacroix~\cite{lacroix} examined several R2V conversion issues, and a method utilizing a preprocessing stage that creates a mask from which edges are eliminated and lines are retained has been suggested. Then clustering is carried out using only the pixels from the mask. In this situation, a novel algorithm called the median shift has been suggested. The labeling procedure that follows should also take into account the type of pixel. The final stage entails a regularization process. In various examples, the significance of the pre-processing ignoring edge pixels while keeping lines has been demonstrated. Additionally, tests demonstrated the superiority of the median shift over both the mean shift and the Vector-Magic clustering method. This paper also showed that better line vectorization can be obtained by enabling the extraction of dark lines, which can support the use of high-pass filters as a preprocessing stage to put further emphasis on those dark lines.

On the straightforward job of denoising additive white Gaussian noise, Xie et al.~\cite{xie} developed a unique strategy that performs on par with conventional linear sparse coding algorithms. In the process of fixing damaged photos, autoencoders are used to lower image noise.

An approach that completes the automatic extraction and vectorization of the road network was presented by Gong et al.~\cite{gong}, first, varied sizes and strong connection; second, complicated backgrounds and occlusions; and third, high resolution and a limited share of roads in the image are the key barriers to extracting roads from remote sensing photos. Road network extraction and vectorization preservation make up the two primary parts of the road vectorization technique in this paper. This study also demonstrates the benefits of dense dilation convolution, indicating the potential for adopting autoencoding models to maintain vectorization.

Fischer and Amesberger~\cite{fischer} showed that preprocessing the raster image with an autoencoder neural network can reduce complexity by over 70\% while keeping a reasonable image quality. They proved that autoencoders perform significantly better compared to PCA in this task. We base our work on this previous work, having a closer look at the effect of high-pass filters on autoencoding in an image vectorization pipeline.

\section{Methodology}
\label{methodology}
In this section, the general approach is described. First, the selected dataset is introduced. The structure of the employed autoencoder is explained next. Details about the software implementation are given, and the processing pipeline is highlighted. Finally, evaluation methods are discussed.

\subsection{CAT Dataset - as Data}
A dataset with over 10,000 cat images is used as the basis for training the autoencoder for evaluating the results. The CAT dataset was published in 2008 by Zhang et al.~\cite{zhang}. The content of the images is secondary for this work: The main reason this dataset is used is the fact that features such as ears, eyes, and noses are relatively easy to see in these images. The autoencoding model can thus be trained on these features and reliably reproduce them.

\subsection{Autoencoder - Functional Structure}
\begin{figure*}
    \centering
  \includegraphics[width=0.8\textwidth]{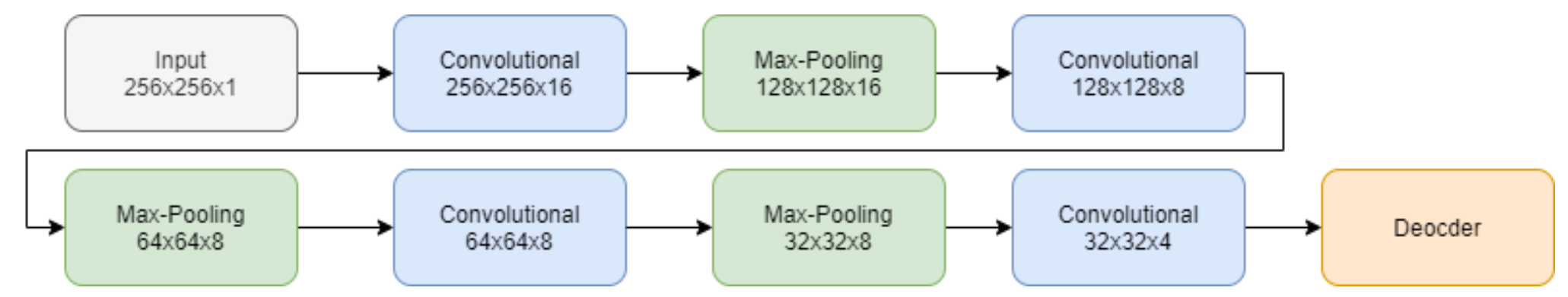}
  \caption{Encoder part of autoencoder}
	\label{fig:encoder-structure}
\end{figure*}

The starting point is input with the size 256 x 256 x 1 (a 256 x 256 grayscale image). The first layer of the autoencoder is a convolution layer that contains 16 different trainable filter kernels. Each kernel can result in a different representation of the input image. A Max-Pooling layer is connected to the convolutional layer to increase the density of the data and reduce the necessary computing power by reducing the number of trainable neurons. This 2x2 layer halved the size of the original image. This convolutional-max-pooling layer cascade is repeated twice for the next two layers, with the convolutional layer having 8 different filters and the same 2x2 max-pooling layer resulting in 64x64 and 32x32 sizes. In the last convolution layer of the encoder, which receives a 32x32 matrix as input, only four convolution kernels are used. The point of highest data density is here reached; therefore, the Max-Pooling layer is omitted. This layer of the autoencoder contains the most compact coding or representation of the data set. Figure~\ref{fig:encoder-structure} shows the encoder part of the autoencoder.

The decoder follows the layer with the highest data density. This part of the autoencoder is responsible for reconstructing the learned encoding. It uses transposed convolution layers and batch normalization layers. The transposed convolution layer works in a similar way to a convolution layer. The difference between the two is that by transposing the input, the layer is no longer compressed but decompressed. Here, the principles of the convolution layer are reversed. The filter kernel is used to determine how the input value is broken down into the larger grid. By using this layer, the image matrix is again enlarged. The transposed convolution layer is followed by a batch normalization layer. These layers, also known as batch norms, serve to accelerate and stabilize the learning process of neural networks. They reduce the amount by which the values of the neurons can shift. On the one hand, the network can train faster because the batch norm ensures that the activation value is neither too high nor too low. On the other hand, using this layer also reduces overfitting since less information is lost through dropouts.

\begin{figure*}
    \centering
  \includegraphics[width=0.8\textwidth]{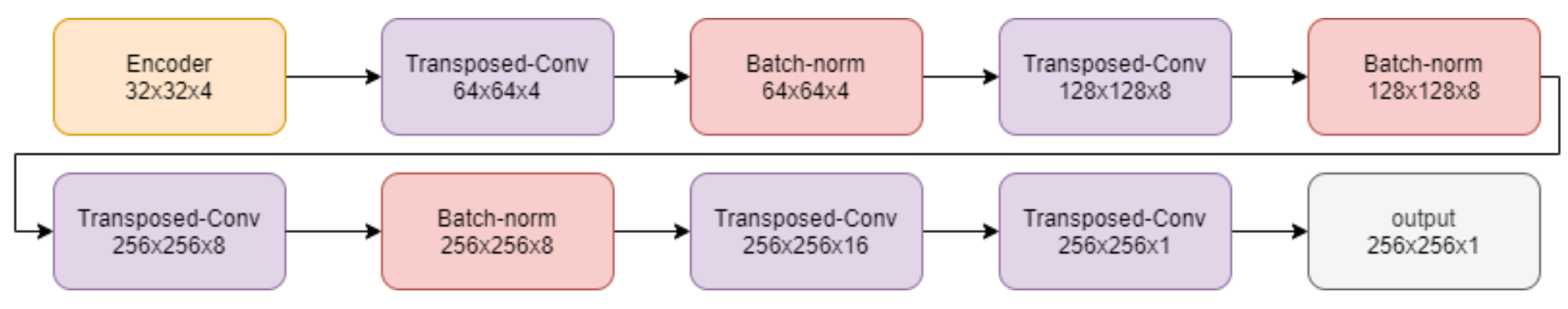}
  \caption{Decoder part of autoencoder}
	\label{fig:decoder-structure}
\end{figure*}

The decoder connects directly to the encoder to take over the most compact representation of the data set passed by the encoding layers. First, the decoder receives a tensor with a size of 32x32x4 as input. The first function that is applied to this tensor is a transposed convolution layer. This results in an enlargement of the image matrix to 64x64. Four 3x3 filter kernels are used here. This is followed by a batch-norm layer to normalize the results and accelerate the learning process. The same process is repeated with a different number of filter kernels to maintain the symmetrical structure of the autoencoder after reaching the original matrix size of 256x256; another transposed convolution layer is added. This ensures that the output of the first layer and the input of the last layer have the same size. The final layer reduces the tensor dimension to one to produce a grayscale image as output.  Figure~\ref{fig:decoder-structure} shows the decoder part of the autoencoder.

\subsection{Software Implementation}
The test/evaluation framework was implemented in Python. The autoencoder was implemented with TensorFlow~\cite{tf} and Keras~\cite{keras}. The convolutional neural network was built with convolution and pooling layers in three steps to a 32×32 bottleneck. The decoder mirrors this structure with three steps of transposed convolutional layers and batch normalization layers. The autoencoder input is set to a 255x255 image (gray-scaled). The high-pass filters used in this paper are the standard implementations in OpenCV~\cite{opencv}. 

\subsection{General Approach of Processing}
Regardless of the path an image takes in any pipeline that will be built, the first processing stage is always going to be converting the image into grayscale. The focus of this work is on single-channel images; however, it can be extended in the future for multi-channel (RGB) processing. Therefore, when a pipeline is demonstrated visually, the initial version of the image displayed is going to be grayscale, but this is implying that the raw RGB images were all grayscaled, which will be a common branch for all the pipelines built in this work.

\begin{figure}
    \centering
    \includegraphics[width=0.95\linewidth]{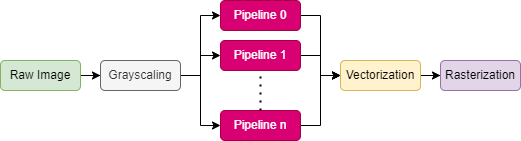}
	\caption{General processing approach}    
	\label{fig:general-processing-approach}
\end{figure}

After an image is grayscaled, it will go through a certain cascade of processing stages. In this paper, the stages concerned are high-pass filtering, autoencoding, and vectorization. The experiments in this work are going to tune the different parameters that these stages can take. More importantly, the outputs of all pipelines possible are going to be in a vector format because we are attempting to enhance the vectorization process while aiming for an abstract representation of the image. Therefore, a rasterization stage is going to always be placed at the end of every pipeline. Converting images back into their raster format is mandatory to perform a comparison between the grayscale image that was initially fed to a pipeline and its resulting vector format. Hence, we rasterize the vector output to be able to evaluate the efficiency of the pipeline. A general processing approach for the different pipelines is shown in Figure~\ref{fig:general-processing-approach}.

\subsection{Evaluation Methods}
The case at hand deals with both vector and raster images. Therefore, for a comparison to take place, a comparison method for each format needs to be selected. 

\begin{itemize}
	\item \textbf{Vector:} Various methods can be used to measure the level of complexity in a vector image. One is the file size, which can be used to calculate the length of all path entries in the file. Furthermore, investigating the reduction of complexity can be done by analyzing the longest path tags. The number of path tags can be taken as a characteristic value of the complexity. In this paper, it is assumed that the number of SVG path entries is directly related to its complexity.
	\item \textbf{Raster:} There are mainly two common ways of comparing raster images. The first one is comparing images based on the mean squared error (MSE)~\cite{mse}. The MSE value denotes the average difference of the pixels all over the image. A higher MSE value designates a greater difference between the original image and the processed image. Nonetheless, it is indispensable to be extremely careful with the edges. A major problem with the MSE is that large differences between the pixel values do not necessarily mean large differences in the content of the images. The Structural Similarity Index (SSIM)~\cite{ssim} is used to account for changes in the structure of the image rather than just the perceived change in pixel values across the entire image. The implementation of the SSIM used is contained in the Python library Scikit-image (also known as "Scikit")~\cite{scikit}. The SSIM method is significantly more complex and computationally intensive than the MSE method, but essentially, the SSIM tries to model the perceived change in the structural information of the image, while the MSE estimates the perceived errors.
\end{itemize} 
In the experiments conducted for this paper, the results of MSE and SSIM drive the same conclusion. Therefore, to avoid redundancy, only the SSIM graphs are displayed in this paper.

\section{Experimentation}
\label{experimentation}
\begin{figure}
    \centering
  \includegraphics[width=0.98\linewidth]{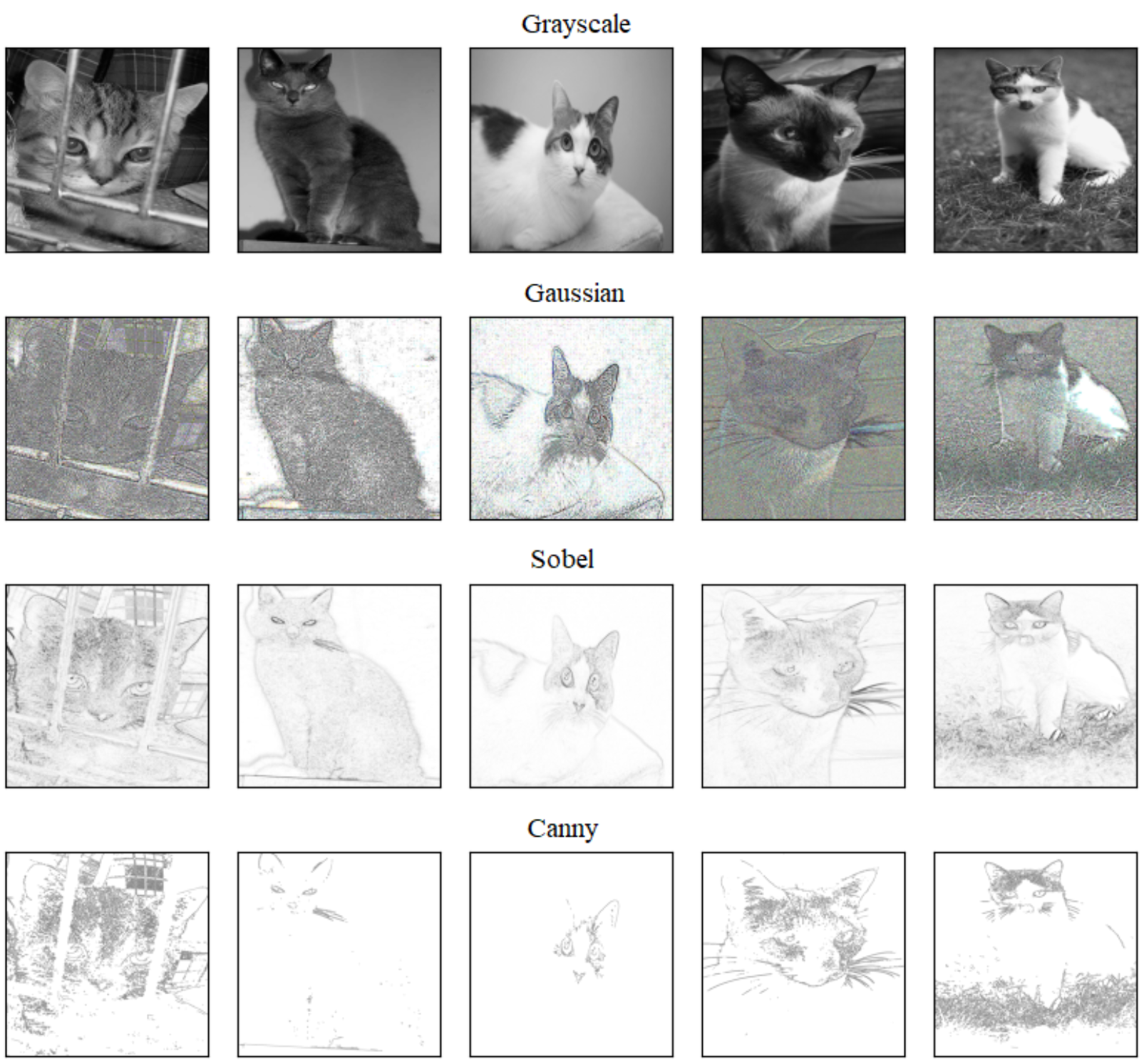}
  \caption{Applying different filters to five random images}
	\label{experiments-first-example}
\end{figure}

Firstly, a sample of five images was filtered with the initial high-pass filters. The results are shown in Figure~\ref{experiments-first-example}.

The first impression is that the Gaussian filter results in some significant noise. Both the Sobel and the Canny filters were acceptable, with the Sobel seemingly having better results for the human eye. Because it made more sense to have the detected lines drawn black on a white image than the opposite case, the three filters were inverted. 

\subsection{Blur-Free Noise-Reduction Filtering}
\begin{figure}
    \centering
  \includegraphics[width=0.98\linewidth]{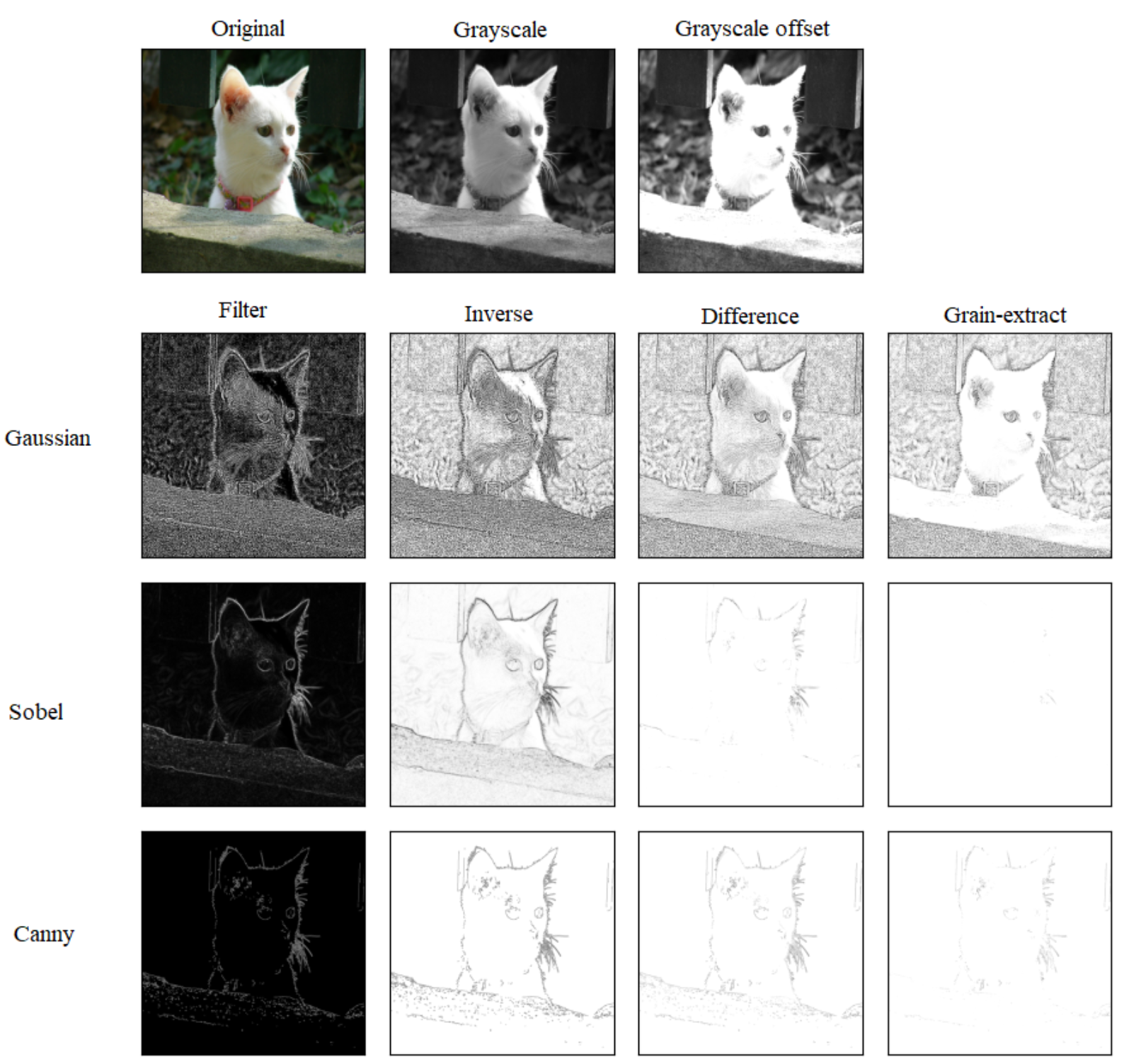}
  \caption{Applying the difference and grain-extract to a random image after being filtered}
	\label{difference-grain}
\end{figure}

In an attempt to reduce the noise the Gaussian filter was causing, two trials were done. They both worked by cascading a filter on top of each high-pass filter. This smoothing filter should result in noise reduction while avoiding blurring the image. Hence, two filters were chosen: difference and grain-extract filters. Figure~\ref{difference-grain} shows the result of applying the two chosen filters on the high-pass filters.

Although the image is still too noisy to be fed into a neural network, the noise-reduction filters may provide a roughly improved version of the Gaussian filter. The difference and grain-extract filters, however, resulted in a decline in image quality and a sizable data loss as compared to the Sobel and Canny filters. The experiment therefore suggests that these two recommended filters are unsuitable for use in a subsequent preprocessing stage and that the Gaussian filter should be categorically excluded from any further use in the project due to its inherent noise.

\subsection{Filter-Inversion Effect on Autoencoding}
The second experiment done in this section is obtaining the difference between training an autoencoder with images whose lines are drawn in black on a white background and training it with the same images but inverted.

\begin{figure}
    \centering
  \includegraphics[width=0.98\linewidth]{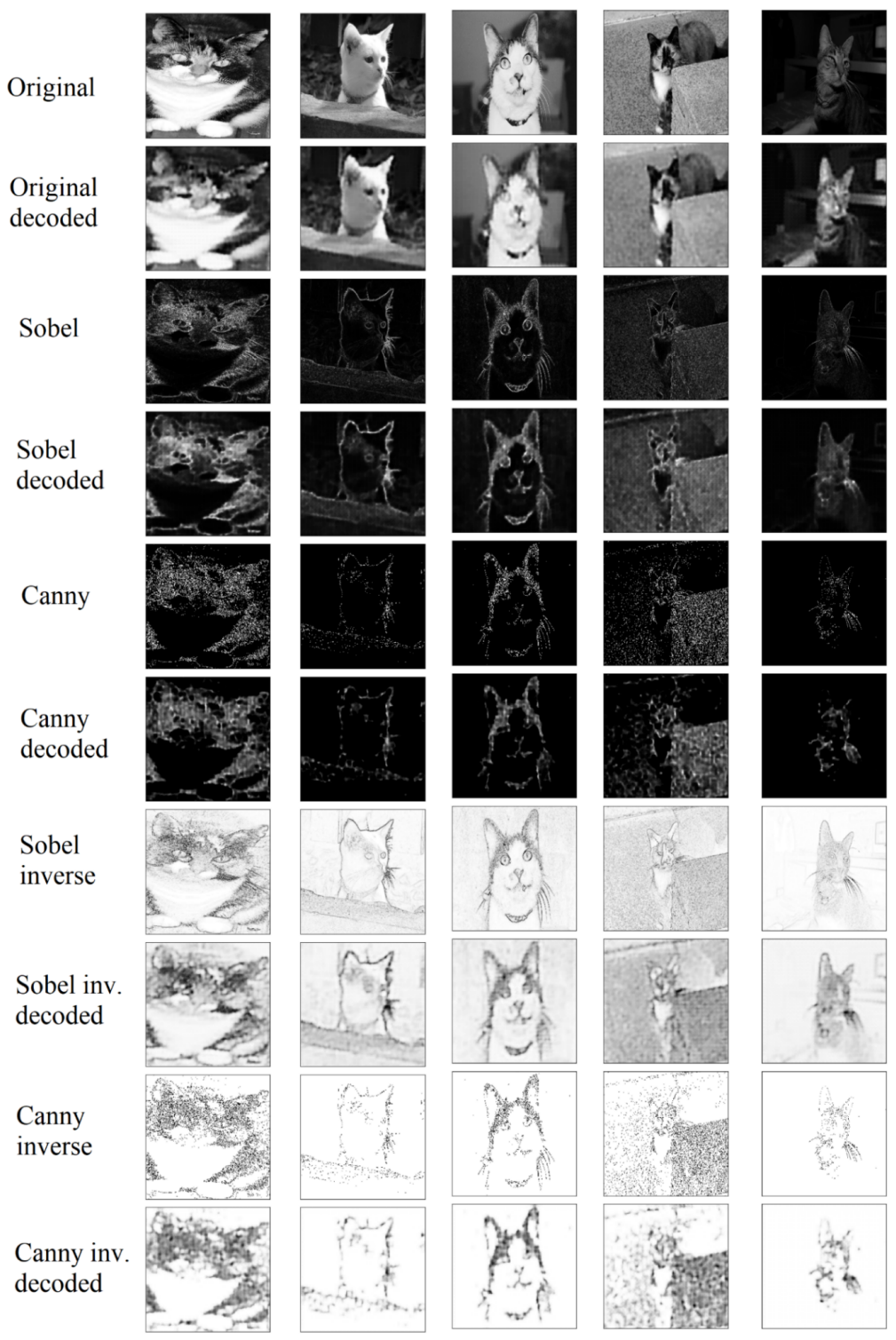}
  \caption{Comparison between the autoencoding of the Sobel and canny filtered images with both of their versions}
	\label{filters-2versions-comparison}
\end{figure}

Therefore, four models of autoencoders were trained with 5000 epochs each in addition to the default model, which makes them five models each trained with the following types of images respectively: grayscale images, Sobel-direct images, Sobel-inverse images, canny-direct images, and canny-inverse images (direct: dark background and white features. inverse: inverse of direct). Five images were selected randomly and put through the five trained models as shown in Figure~\ref{filters-2versions-comparison}.

\begin{figure}
    \centering
  \includegraphics[width=0.98\linewidth]{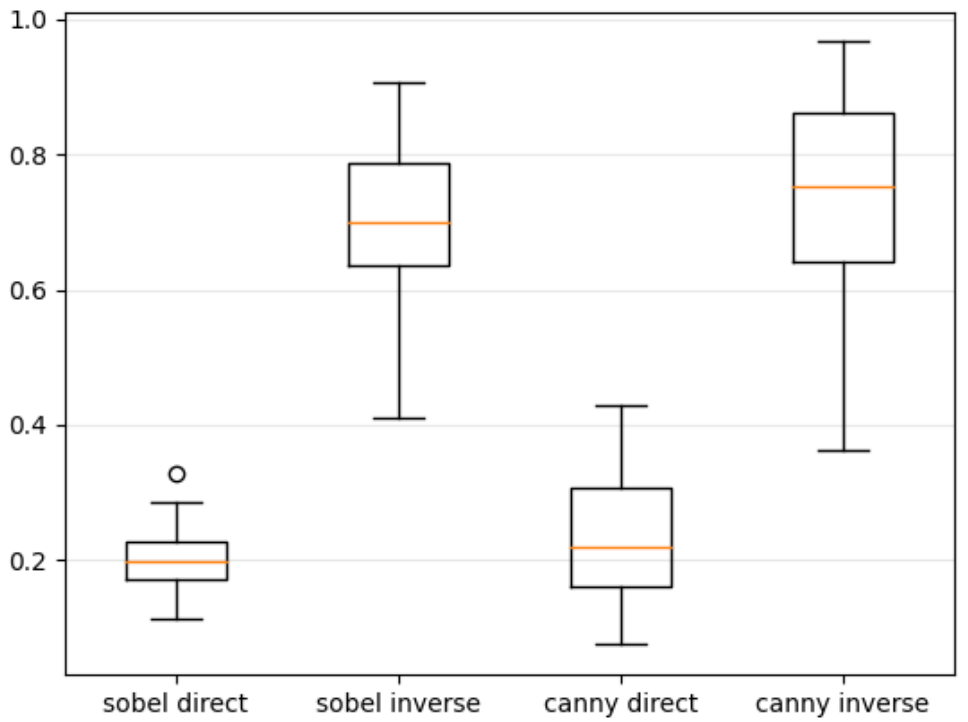}
  \caption{SSIM of different autoencoding approaches}
	\label{ssim-different-autoencoding-approaches}
\end{figure}

The first conclusion drawn was that, when training an autoencoder, the semi-supervised neural network responds better when the training images have darker lines in their important features. However, a rough estimation with the human eye would not do, but rather an exact mathematical calculation. Therefore, a measurement of similarity was done between every image and its decoded version. This was a better way of using the SSIM than comparing them with the default images, as the goal was to determine how close the autoencoding was. For this part, 50 images were used to dampen the image-specific features and make the measurement more generalized.The measured values were plotted in Figure~\ref{ssim-different-autoencoding-approaches}.

For the sobel-direct, the mean and standard deviation values were 0.202 and 0.044, respectively. Their inverse scores were 0.699 and 0.124, respectively. For the canny-direct, the mean and standard deviation values were 0.234 and 0.090, respectively. The inverse scored 0.741 and 0.150, respectively.

These values support our first observation, which is that the autoencoder learns faster when the image's most important features are darker than the rest of the data. The experiments so far have resolved into using the Sobel and Canny filters, and more specifically, their inverted results. At the start, it was thought that the experiments would resolve into choosing only one filter as a preprocessing stage for the autoencoding, but as calculated previously, the quality of images between the Sobel and Canny images is so close that it does not imply the disregard of one of the two filters.

Nevertheless, there is a significant drop in quality when applying a high-pass filter to the original image and then passing it through an autoencoding stage. This raised a flag that perhaps the pipeline's order might not be thorough. For instance, the autoencoder is perceived to work as a reconstruction algorithm. Simultaneously, it can be considered to smooth the image, or in other words, to represent it with more coherence between the pixel values. As a result, the high-pass filters may be more efficient if applied after image reconstruction rather than before autoencoding, which appears to cancel out some of the emphasis generated by the filters. Hence, an experiment on the matter should be performed.

\subsection{Autoencoders as a Preprocessing Stage to High-Pass Filters} 
\begin{figure}
    \centering
  \includegraphics[width=0.98\linewidth]{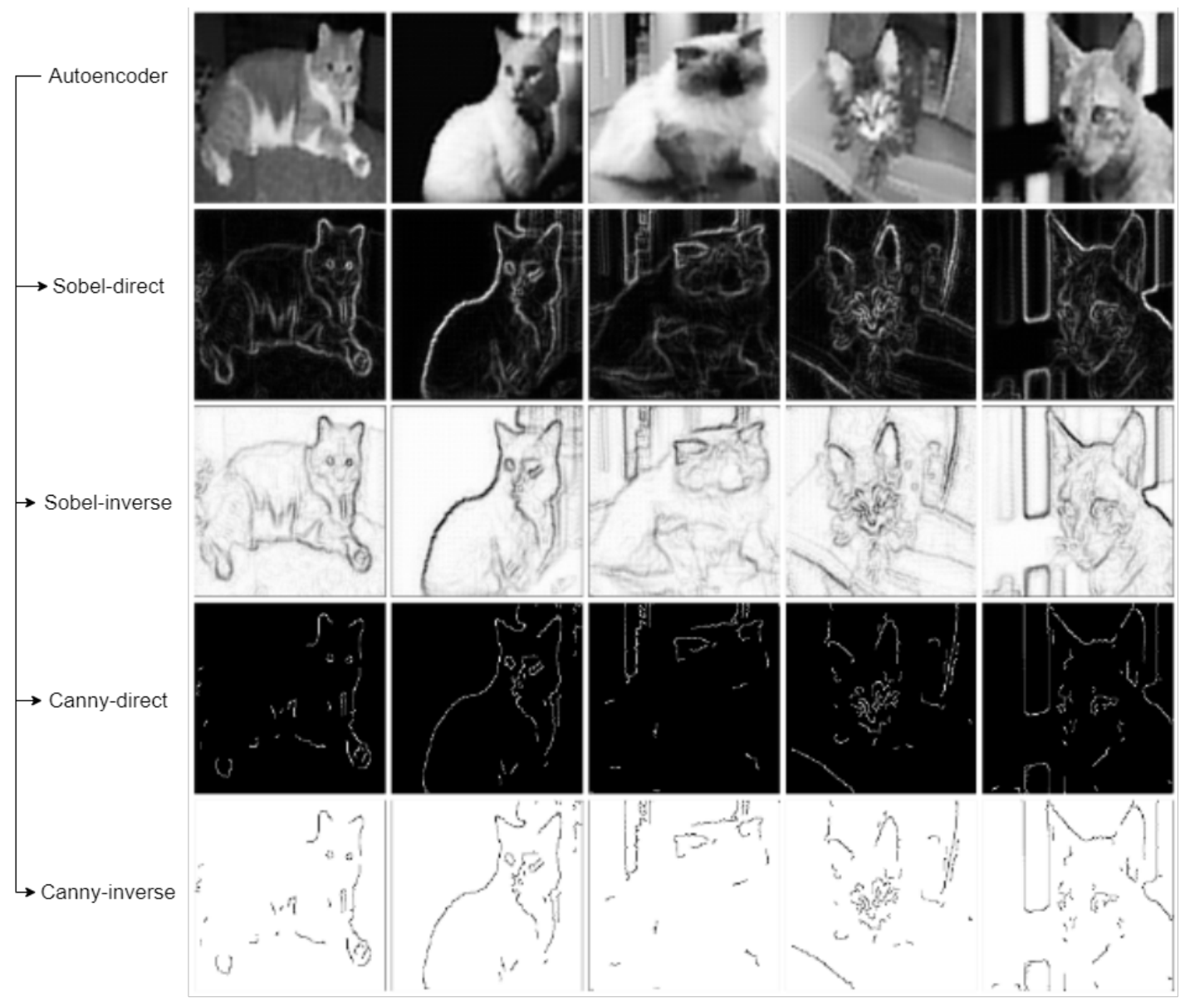}
  \caption{Filtered autoencoder images with Sobel and canny (both versions each)}
	\label{experiment-autoencoder-filters}
\end{figure}

In this experiment, random images were taken, reconstructed with an autoencoder, filtered, and then vectorized. This experiment aims to display the effect of high-pass filters on reconstructed image vectorization. The five random resulting images are shown in Figure~\ref{experiment-autoencoder-filters}.

The first impression the experiment gives off is that the filters brought more definition to the lines in the images, which made the shapes appear clearer. This can lead to better vectorization, as it depends on the definitions of the shapes represented in the tags.

\begin{figure}
    \centering
  \includegraphics[width=0.98\linewidth]{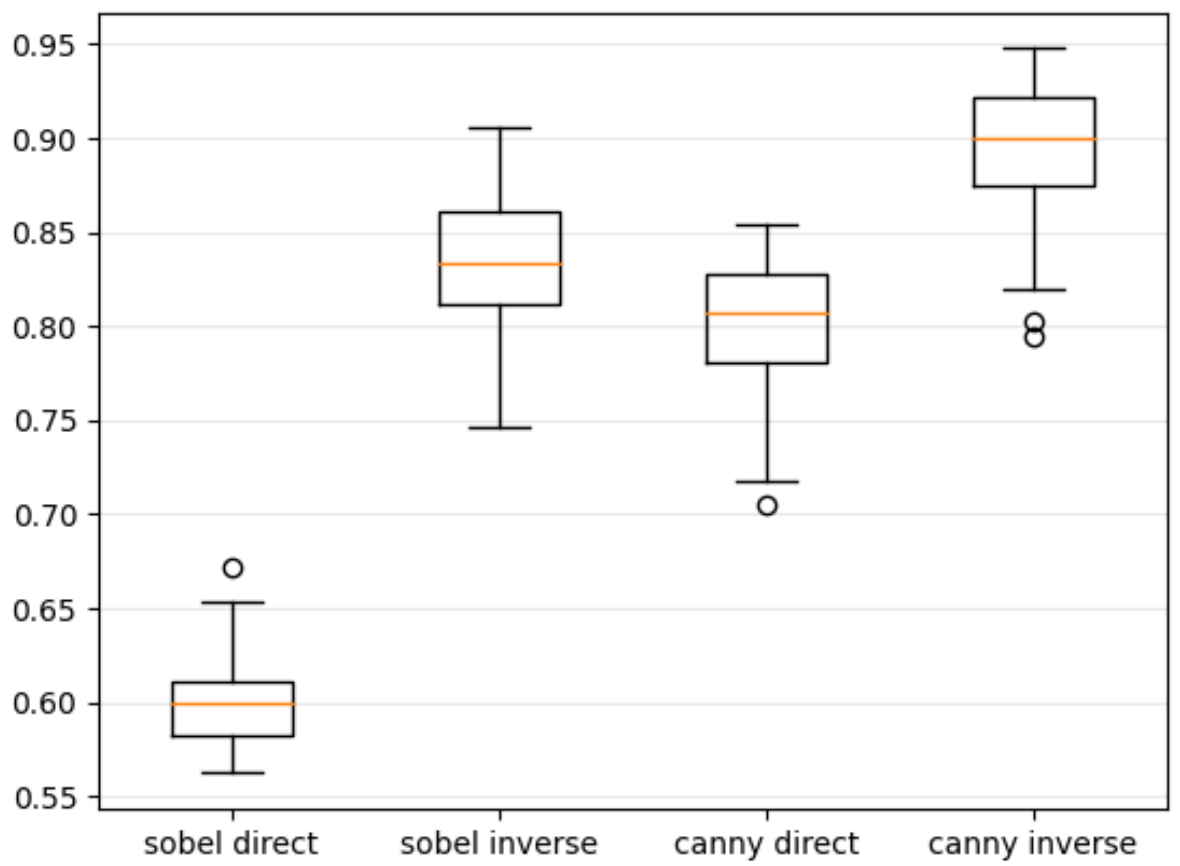}
  \caption{SSIM comparison of the vectorization of each of the four groups of images}
	\label{experiment-ssim-vect}
\end{figure}

However, there are two versions of each of the two filters, which suggest an evaluation of the vectorization of each of the four result groups. Therefore, an SSIM calculation was done between every filtered image and its vector format in a pool of 50 images, randomly selected. The results are displayed in Figure~\ref{experiment-ssim-vect}.

The box plots show the better fitness of white images with black lines when compared to the darker images in vectorization. Visually speaking, the Sobel filter results were more recognizable to the naked eye. However, it left more complexity in the image, which made it harder for the vectorization to be more exact. Therefore, it is concluded that the darker shapes are going to be used in both filters, while there is not yet a clear endpoint to resolve depending on only one of the two filters. Hence, a parallel stage of execution is introduced, which takes the autoencoder images and filters them with one filter before passing them to the global vectorization stage.
 

\section{Evaluation}
\label{evaluation}

Evaluation is concerned with how abstract the resulting images are.  As there are two pre-processing blocks (filtering and autoencoding), four different pipelines can be built: autoencoding, filtering, autoencoding-filtering, and filtering autoencoding. After one of these selections is fed the images, a vectorization process is always cascaded at the end.

First, all of the resulting images are going to be evaluated based on their path count (size) and similarity to the input images. Then, a summary of the evaluation is going to be introduced for each of the pipelines individually.

Before engaging in the evaluation, it is good to elaborate on the column naming of the upcoming plots:
\begin{itemize}
	\item default: the default image.
	\item Sobel, canny: the filtered version of the image by the respective filter.
	\item dec: the decoded version.
	\item vect: the vectorized version.
	\item A combination of two or more indicates the case of cascaded stages. A default-dec-sobel label represents the following: the default image is reconstructed with the autoencoder and then filtered with the sobel filter.
\end{itemize}

\subsection{Evaluating the size of the produced images}
\begin{figure}
    \centering
    \includegraphics[width=0.4\textwidth]{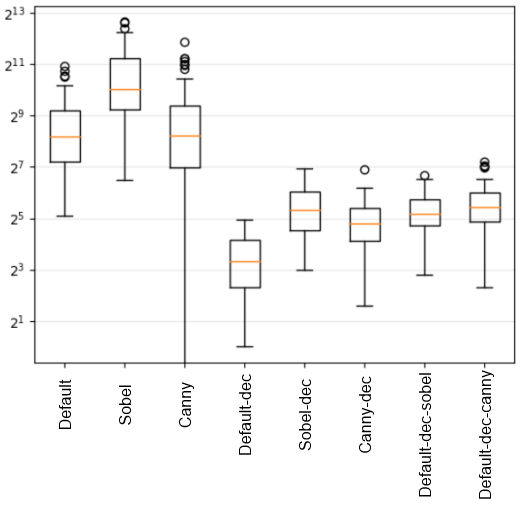}
	\caption{Path count of the resulted groups of vector images}    
	\label{fig:path-count}
\end{figure}

To evaluate the size of the image, we count the number of path objects generated in the SVG file. From Figure~\ref{fig:path-count} (note that the graph is in logarithmic scale) we see that the autoencoder (*-dec-*) significantly reduced the size of images, as it keeps only the most important features. The reconstructed filtered images (canny-dec, sobel-dec) had a similar path count. Although it was much smaller than the ones that did not go through that step, it was still above the default images that were reconstructed and vectorized without any filtering. Finally, when filters were applied to the default images that were put through an autoencoding stage (default-dec-sobel, default-dec-canny), these images scored in size calculations very similarly to the filtered images when only reconstructed (canny-dec, sobel-dec).

\subsection{Evaluating the quality of the produced images}
\begin{figure}
    \centering
    \includegraphics[width=0.4\textwidth]{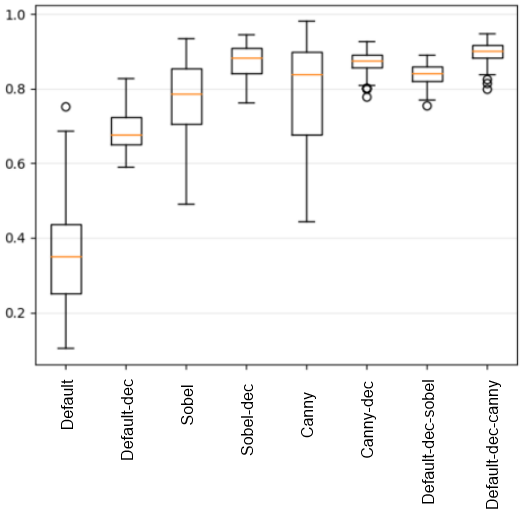}
	\caption{Vectorization accuracy of different pipelines}    
	\label{fig:vect-accuracy}
\end{figure}

A more accurate way of examining the efficiency of the vectorization process of each pipeline is to compare the images and their vector versions (Figure~\ref{fig:vect-accuracy}). The pipeline of autoencoding-filtering-vectorization (two last groups on the most-right) seems to experience the highest SSIM, which indicates its fitness in vectorization. It made more sense for the autoencoder to reconstruct the images and then for the filters to come afterward, emphasizing the important features of each image.


\subsection{Implemented Pipelines: an evaluation summary}
This is a summary of the evaluation of the results for each of the pipelines individually.
\begin{itemize}
	\item \textbf{Autoencoding-Vectorization:} This pipeline was based on the work of Fischer and Amesberger~\cite{fischer}. However, the implementation was different, and the evaluation was about the abstractness of the results. The quality of the vectorization is acceptable only in terms of general similarity. However, an abstract representation of the image is not achieved (Figure~\ref{fig:dec-vect}).
	\begin{figure}
	    \centering
	    \includegraphics[width=0.9\linewidth]{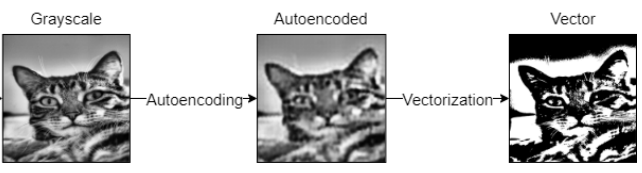}    
		\caption{Autoencoding-vectorization pipeline}    
		\label{fig:dec-vect}    
	\end{figure}
	
	\item \textbf{Filtering-Vectorization:} In this pipeline (Figure~\ref{fig:filter-vect}), the vectorization algorithm finds difficulty in vectorizing the filtered images. This is due to the noises caused by the applied filters. Although the experiments showed that the quality of the vectorization increased when the images were taken as a light background with dark features, the noise involved created an obstacle for Potrace to convert thoroughly the images into a vector format, which resulted in losing data.
	\begin{figure}
	    \centering
	    \includegraphics[width=0.9\linewidth]{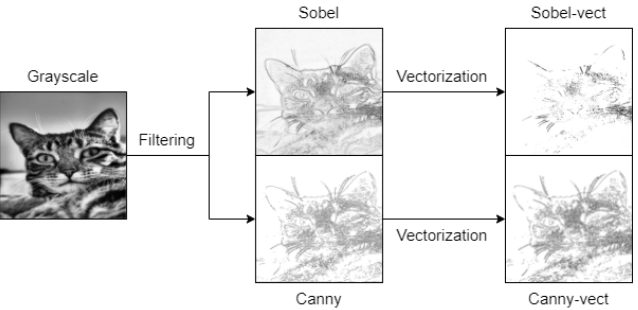}    
		\caption{Filtering-vectorization pipeline}    
		\label{fig:filter-vect}    
	\end{figure}
	
	\item \textbf{Filtering-Autoencoding-Vectorization:} This pipeline was built as an attempt to enhance the \textit{Autoencoding-Vectorization} pipeline. Although the autoencoding stage was efficient in reducing the size of the images, it did not result in an abstract view of the image features. Therefore, a filtering stage was placed before the autoencoding process. Unfortunately, this pipeline does not achieve the result intended. The autoencoding stage was supposed to reconstruct the filtered images in a lower complexity; but the case at hand is that the autoencoding model is attempting to smooth the images, canceling the effect of the high-pass filters. This has resulted in a significant drop in the quality of the vector images, which is seen in Figure~\ref{fig:filter-dec-vect}.
	\begin{figure}
	    \centering
	    \includegraphics[width=\linewidth]{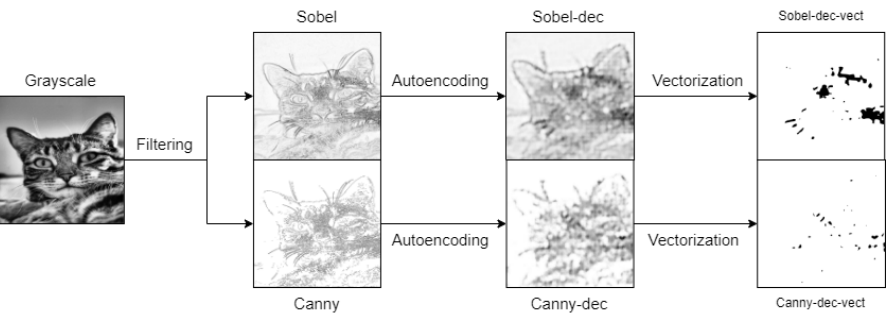}    
		\caption{Filtering-autoencoding-vectorization pipeline}    
		\label{fig:filter-dec-vect}    
	\end{figure}

	\item \textbf{Autoencoding-Filtering-Vectorization:} Due to the results in the \textit{Filtering-Autoencoding-Vectorization} pipeline, it was clear that the filtering stage would act more appropriately if it succeeded the autoencoding process, rather than preceding it. This was concluded when the autoencoding model was seen to reduce the complexity of the images while introducing a smoothing effect. The filters were placed after the reconstruction stage to preserve the important features of the reduced-complexity image. This cascade shows an acceptable vectorization quality while resulting in the intended abstract representation of the images as shown in Figure~\ref{fig:dec-filter-vect}. \\
	\begin{figure}
	    \centering
	    \includegraphics[width=0.5\textwidth]{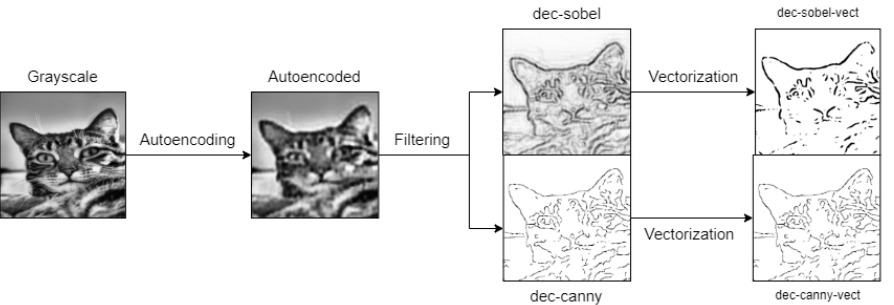}    
		\caption{Autoencoding-filtering-vectorization pipeline}    
		\label{fig:dec-filter-vect}    
	\end{figure}
As for providing more visualizations of the results that can be obtained with this pipeline, Figure~\ref{fig:input-output} shows some random images that were fed to the Autoenconding-filtering-vectorization pipeline along with their respective output images. As can be seen, the features of the cats are extracted very clearly in all examples.
	\begin{figure}
	    \centering
	    \includegraphics[width=0.99\linewidth]{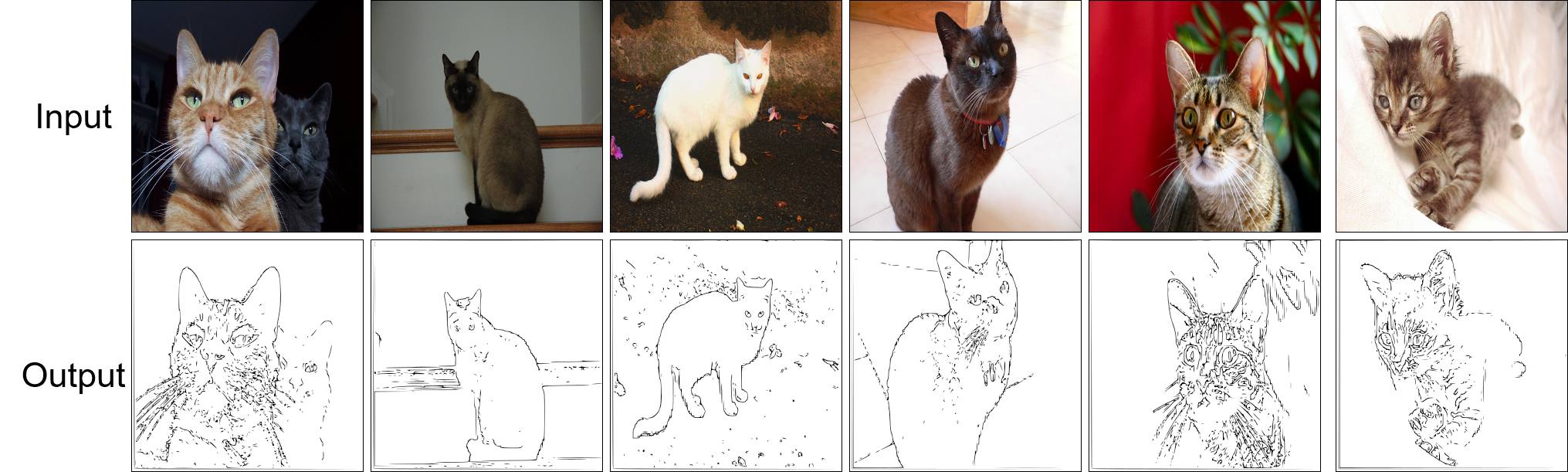}    
		\caption{Some of the output images along with their input images of the pipeline Autoenconding-filtering-vectorization}    
		\label{fig:input-output}    
	\end{figure}

\end{itemize}

\section{Conclusion}
\label{conclusion}

This paper discusses the autoencoding step and the use of high-pass filters in vectorization pipelines. As demonstrated, high-pass filters can improve the training of an autoencoder, which in turn improves the efficiency of vectorization by maintaining key aspects of an image.

The images that underwent the cascade of autoencoding-filtering scored the greatest in similarity and the lowest in error after the vectorization algorithm's effectiveness in each pipeline was assessed. This indicates that the most crucial elements of the reconstructed images were maintained and that the filtering step that came after the reconstruction enhanced those features even further, resulting in a better vectorization and a more abstract representation of the image.

Although the results from this cascade of autoencoding-filtering were respectable and met the initial objectives, more work needs to be done on the training dataset and model structures.

Regarding future work, experiments showed that dark features on a light background in images can improve both the training of autoencoder models and the process of vectorization. This will be an issue for further investigation. As this paper deals with single-channel (i.e., gray-scale) images, another aspect of the investigation will be the vectorization of multi-channel images.

\bibliographystyle{IEEEtran}
\bibliography{references}

\end{document}